\documentclass{article}
\usepackage[utf8]{inputenc}
\usepackage[round]{natbib}
\usepackage{bibentry}
\usepackage[hyphens]{url}
\usepackage{float}
\usepackage{amsfonts}
\usepackage[]{algpseudocode}
\usepackage{algorithm}
\usepackage{graphicx}
\usepackage{multirow}
\usepackage{makecell}

\usepackage[colorlinks = true,
            linkcolor = blue,
            urlcolor  = red,
            citecolor = blue,
            anchorcolor = blue]{hyperref}

\usepackage{doi}

\newcommand{\hp}[2] {{#1}_{\mathrm{#2}}}

\begin{document}

\title{Graph Node Embeddings using Domain-Aware Biased Random Walks}
\author{Sourav Mukherjee \thanks{sourav.mukherjee.8@gmail.com}
\and Tim Oates \thanks{tim.oates@synaptiq.ai}
\and Ryan Wright \thanks{ryan@galois.com}}
\date{}
\maketitle

\begin{abstract}

The recent proliferation of publicly available graph-structured data has sparked an interest in machine learning algorithms for graph data. Since most traditional machine learning algorithms assume data to be tabular, embedding algorithms for mapping graph data to real-valued vector spaces has become an active area of research. Existing graph embedding approaches are based purely on structural information and ignore any semantic information from the underlying domain. In this paper, we demonstrate that semantic information can play a useful role in computing graph embeddings. Specifically, we present a framework for devising embedding strategies aware of domain-specific interpretations of graph nodes and edges, and use knowledge of downstream machine learning tasks to identify relevant graph substructures. Using two real-life domains, we show that our framework yields embeddings that are simple to implement and yet achieve equal or greater accuracy in machine learning tasks compared to domain independent approaches. 

\end{abstract}

\section{Introduction}
\label{sec:Introduction}

In recent years, we have witnessed a dramatic increase in the volume of graph-structured data being generated and made publicly available. For example, the Stanford Large Network Dataset Collection\footnote{\url{https://snap.stanford.edu/data/}} provides graph data from a variety of domains such as social networks, web graphs (where nodes and edges represent web pages and hyperlinks, respectively), citation and collaboration networks, and so on \citep{snapnets, biosnapnets}. Linked Data \citep{Bizer2009LinkedD} makes it possible for open data providers to publish graph structured datasets using a standardized representation, namely, the Resource Description Format (RDF) \citep{DBLP:conf/semweb/SchmachtenbergBP14}. The standardized representation allows graph data from heterogeneous sources to be combined into a unified, web-scale graph, namely, the Linked Open Data (LOD) Cloud\footnote{\url{https://lod-cloud.net/}}, that may be searched, crawled and indexed just like the world wide web. These developments have motivated a strong interest in machine learning algorithms for extracting insights from graph data.

Traditional machine learning methods frequently rely on tabular data representation. Specifically, the existence of real-valued vector representations of instances, and of a real-valued distance metric between pairs of instances, are assumed. Consequently, such approaches are not readily applicable to graph data. However, it is possible to overcome this limitation by defining a  mapping from entities in graph-structured data to points in a real-valued vector space $\mathbb{R}^d$. Discovering such mappings, also known as \emph{embeddings}, has emerged as an active area of research. A desirable embedding should: (a) be easy to compute, (b) map to a space of low dimensionality, and (c) induce a distance metric consistent with the notion of similarity in the original domain. Vector representations resulting from such an embedding may then be used to perform a variety of downstream machine learning tasks such as supervised learning (classification and regression), entity and document modeling, recommendation systems, etc., as demonstrated by  \citet{DBLP:conf/semweb/RistoskiP16}.

A common approach to graph embedding draws inspiration from a language modeling technique, namely word2vec  \citep{DBLP:journals/corr/abs-1301-3781}, that maps words in natural language to real-valued vectors based on their occurrences in sentences of a text corpus. Many graph embedding approaches perform random walks on graphs to generate linear sequences, which are treated as sentences and fed as input to word2vec to produce vector representations \citep{DBLP:conf/kdd/PerozziAS14, DBLP:conf/kdd/GroverL16, DBLP:conf/semweb/RistoskiP16}. 

The random walk strategies used in the above approaches rely only on graph structure. Any semantic information from the underlying graph domain is ignored. Further, no knowledge of intended downstream applications is assumed. In this paper, we argue that semantic information from underlying domains and knowledge of downstream tasks can help improve embeddings. As an illustrative example, consider the problem of targeted advertising using a social network graph. When the nature of the advertised product is known, we are able to identify some types of graph substructures as  more relevant than others. If, on the other hand, the goal is to identify terrorist networks, then a different set of substructures may deserve greater attention. Therefore, we present biased random walk strategies informed by domain knowledge such as domain-specific interpretations of graph nodes and edges, and knowledge of downstream machine learning tasks. We show that such strategies are simple to define and implement, and can lead to embeddings that lead to equal or higher accuracies in machine learning tasks compared to domain independent approaches.   

The rest of this paper is organized as follows. Section \ref{sec:RelatedWork} reviews the literature.  Section \ref{sec:Methodology} presents our methodology. Section \ref{sec:Evaluation} describes our experimental setup and presents results comparing our methods to existing approaches. Finally, Section \ref{sec:Conclusion} concludes the paper.

\section{Related Work}
\label{sec:RelatedWork}

The terminology used in machine learning for graphs  is sometimes unclear in conveying whether the instances in a dataset are nodes of the same graph or whether each instance is a graph itself. For example, the term ``graph clustering'' can refer to the task of clustering the nodes of a graph based on their inter-connections, as well as to the problem of clustering a set of objects where each object is itself a graph. Deep learning terminology for graph data can be similarly ambiguous, as can be seen, for example, in the application of convolutional networks to graph data. \citet{DBLP:journals/corr/HenaffBL15} apply such networks to data represented as a single graph whose nodes represent instances and  edges represent a non-Euclidean distance metric. On the other hand, \citet{DBLP:conf/icml/NiepertAK16} use convolutional networks on datasets where every instance is a graph. Another example is the unsupervised learning problem of generating embeddings, where the goal is to map objects to dense vector representations in a low dimensional Euclidean space. In the work by \citet{DBLP:conf/kdd/PerozziAS14}, the instances being embedded are nodes of the same graph (e.g., users in an online  network), while \citet{DBLP:conf/kdd/YanardagV15} consider datasets where each instance is itself a graph (e.g., protein structures).

To distinguish between these two types of scenarios, we use the following terminology in relation to the task of learning a function $f$ from a set of training instances (labeled or otherwise) $\{\mathbf{x}_i\}_{i=1}^N$. 
\begin{itemize}
\item We use  \emph{graph mining} to refer to scenarios where  the instances $\{\mathbf{x}_i\}_{i=1}^N$ are nodes of the same graph. As observed by \citet{DBLP:conf/kdd/PerozziAS14}, the training instances in graph mining are not independent and identically distributed (i.i.d.), since the graph edges denote relationships between the instances. Therefore, graph mining amounts to relational learning. Real-world examples of graph mining include:
\begin{itemize}
\item Community detection in social networks, where the users (nodes) in a social network graph are partitioned based on links between them.
\item Classifying research publications by subjects or keywords from a citation network of papers. 
\end{itemize}
\item We use \emph{learning from graph databases} to refer to scenarios where each instance $\mathbf{x}$ is itself a graph, and the training set $\{\mathbf{x}_i\}_{i=1}^N$ consists of i.i.d. samples from an underlying distribution of graphs. Real examples include:
\begin{itemize}
\item Classifying chemical compounds as carcinogenic or non-carcinogenic.
\item Classifying proteins as enzymes or non-enzymes.
\end{itemize}
\end{itemize}

In the past five years, both \emph{graph mining} and \emph{learning from graph databases} have been explored by deep learning researchers; however, graph mining has been studied much more extensively. Our present work falls under the category of graph mining.

In unsupervised graph mining, the problem of generating \emph{node embeddings} (also known as \emph{graph embeddings}) has gained a lot of attention. A node embedding is a mapping of the nodes in a graph to points in a low-dimensional Euclidean space. Intuitively, the goal is to have ``similar'' nodes in the graph mapped to nearby points in the Euclidean space. Once generated, the same embeddings may be used in a wide range of machine learning tasks such as supervised learning (classification and regression), entity and document modeling, recommendation systems, etc., as demonstrated by  \citet{DBLP:conf/semweb/RistoskiP16}.

A seminal contribution in this field is the DeepWalk algorithm introduced by \citet{DBLP:conf/kdd/PerozziAS14}, which applies language modeling techniques to graph mining. DeepWalk starts by generating truncated random walks on the graph. These random walks provide an approximate representation of  neighborhood structures in the graph. The generated walks are then treated as sentences on which word embedding techniques such as Skip-gram word2vec \citep{DBLP:journals/corr/abs-1301-3781} are trained to generate dense vector representations of nodes. The authors validated DeepWalk using multi-label classification in BlogCatalog, Flickr, and YouTube user networks.  

Subsequently, this approach has been extended in various ways.
Most extensions of DeepWalk redefine how neighborhood structures are represented. For instance, \citet{DBLP:conf/kdd/GroverL16} introduce the concept of $2^{\mathrm{nd}}$ order random walks in their algorithm, node2vec. In a $2^{\mathrm{nd}}$ order random walk, two parameters $p$ and $q$ determine the likelihood of returning to the node visited just prior to the current node, and of moving further away from that previous node, respectively. Various settings of these parameters lead to walks biased in various ways. This approach was validated using multi-label classification in BlogCatalog, Protein-Protein Interaction (PPI), and Wikipedia graph datasets. 

Another extension of DeepWalk is the RDF2Vec algorithm by \citet{DBLP:conf/semweb/RistoskiP16}. RDF\footnote{\url{https://www.w3.org/RDF/}} is a standardized format for Linked Open Data (LOD) \citep{DBLP:conf/semweb/SchmachtenbergBP14}, and has been used to represent several web-scale directed graphs such as DBPedia, Wikidata, etc. RDF2Vec generates node embeddings for such graphs. It employs various types of  neighborhood representations, viz. (a) enumerating \emph{all} walks of a given depth rooted at every node generated using breadth-first search, (b) Weisfeiler-Lehman Subtree RDF Graph Kernels  \citep{DBLP:journals/ws/VriesR15}, and (c) a given number of random walks of up to a given depth rooted at every node. However, for web-scale datasets, only random walks are used because  computing the other representations becomes prohibitively expensive.
\citet{DBLP:conf/wims/CochezRPP17} improve this method by biasing the random walks using various strategies (in a biased random walk, all outgoing edges originating from a node are not equally likely to be selected for a hop from that node). Examples of biasing strategies include \emph{predicate frequency} where the probability of traversing a (labeled) edge is proportional to the frequency of occurrence of its edge label in the graph, \emph{object frequency} where the probability is proportional to the in-degree of the node to which the edge leads, and so on. In every case, the biasing strategy is expressed as an edge weighting function that assigns a non-negative weight to each edge. Weighted random walks are then performed by selecting edges with probabilities proportional to the weights assigned to them by the edge weighting function.

RDF2Vec may also be viewed as an extension of DeepWalk from the standpoint of node heterogeneity. The graphs on which DeepWalk was validated are characterized by homogeneous node types (e.g., every node in the YouTube user network represents a human user). RDF graphs such as DBPedia, on the other hand, inherently allow nodes of different types to be present in the same graph. Thus, RDF2Vec embeds nodes of various types into the same Euclidean space. Another approach to embedding nodes of different types (such as image and text) into the same space is the Heterogeneous Network Embedding (HNE) architecture by \citet{DBLP:conf/kdd/ChangHTQAH15}. While DeepWalk, node2vec, and RDF2Vec rely on word2vec (which uses a shallow network with a single hidden layer), HNE uses a multilayer neural network. Conceptually, the input to the network is a pair of nodes, and the output is a predicted similarity metric. 
To accommodate heterogeneous nodes,  every possible pair of node types (e.g., image-image, image-text, text-text, etc.) has a corresponding hierarchical feature extractor module whose inputs are pairs of nodes.  The outputs of these modules, i.e., the extracted features, are  fed into a common prediction layer that predicts similarity between the nodes, which is then used for backpropagation. The backpropagation, in turn, leads to learning the weights that define the embedding function.

All embedding methods described above assume the underlying graph (whether directed or undirected) to be unweighted. \citet{DBLP:conf/aaai/CaoLX16} extend node embeddings to weighted graphs; in their method, structural information is represented using a random surfing model as opposed to random walks.  

Our work is most closely related to the Biased RDF2Vec approach by \citet{DBLP:conf/wims/CochezRPP17}. While \citeauthor{DBLP:conf/wims/CochezRPP17} have shown that biasing random walks can improve the quality of the resulting node embeddings (and the accuracy of downstream machine learning tasks) compared to uniform random walks, the biasing strategies they use are purely based on structural information from the graph; all semantic information is ignored. We propose an alternate approach that utilizes this semantic information. Specifically, we make the following contributions:
\begin{enumerate}
\item We use knowledge of  the underlying graph domain (e.g., ``What do the nodes and edges of this graph represent?'') to formulate random walk biasing strategies that are domain specific. 
\item We show that the domain-aware strategies presented in this paper do not require graph-wide computations such as the frequency of each edge label in the graph, or the in-degrees of every node with an inbound edge, that domain-independent strategies depend on.
In terms of computational cost, this implies that our methods don't require graph-wide queries, and rely solely on local neighborhood information which is amenable to caching due to locality of reference.
\item We use real world data to show that appropriately selected domain-specific biasing strategies produce  embeddings with either comparable or substantially improved accuracy (depending on the underlying graph) in downstream machine learning tasks such as classification.
\end{enumerate}

The next section describes our contributions in detail.

\section{Methodology}
\label{sec:Methodology}

We begin this section by reviewing existing methods that serve as building blocks in our approach. We then present our novel approach of incorporating semantics from  underlying graph domains to generate high quality node embeddings.

\subsection{Preliminaries}
\label{subsec:Preliminaries}

\subsubsection{Word2vec}
\label{subsubsec:Word2vec}

Word2vec, by \citet{DBLP:journals/corr/abs-1301-3781}, is a natural language processing (NLP) technique that scans a corpus of text and produces real-valued vector representations  (also known as a \emph{word embeddings}) of all the unique words in that corpus\footnote{with the exception of ``stop words'' such as ``a'', ``an'', ``the'' etc.}. The goal of embedding is to have similar words mapped to nearby vectors in the Euclidean space containing them, where similarity roughly implies occurrence in similar contexts. 

\citeauthor{DBLP:journals/corr/abs-1301-3781} give two alternative neural architectures for computing word embeddings, namely, Continuous Bag of Words (CBOW) and Skip-gram. Both architectures are \emph{shallow}, in that each has only one hidden layer.  The two architectures differ in how the relationship between a word and its contexts is viewed. In CBOW, the network accepts a context of words  as input, and computes, for every word in the vocabulary, the probability of that word occurring at the center of that context. In Skip-gram, the input to the network is a single word, and the output is a probability mass function over the vocabulary that gives the probability of finding each word as a result of randomly selecting a word within a certain window around that input \citep{McCormick2016}.

These word embedding architectures require data to be sequential (e.g., sentences of words).  However, they have been applied to higher dimensional data, namely graphs, by extending the notion of context from a one-dimensional window around a word to a neighborhood structure of a graph's node. A set of walks rooted at any node may be considered an approximate representation of the node's neighborhood. The advantage of such a representation is that it is sequential, and therefore may be ingested by a word embedding architecture to produce embeddings of graph nodes.
Although this approach has been utilized in a number of algorithms (e.g., DeepWalk by \citet{DBLP:conf/kdd/PerozziAS14}, node2vec by \citet{DBLP:conf/kdd/GroverL16}, and RDF2Vec by \citet{DBLP:conf/semweb/RistoskiP16}, here we focus on the RDF2Vec algorithm due to its relevance to the present work.

\subsubsection{RDF2Vec}
\label{subsubsec:RDF2Vec}

RDF2Vec uses several techniques to represent neighborhood structures as linear sequences: 
(a) enumerating all walks of a given depth rooted at every node, generated using breadth-first search, (b) Weisfeiler-Lehman Subtree RDF Graph Kernels  \citep{DBLP:journals/ws/VriesR15}, and (c) a set of random walks rooted at every node, where the maximum number of walks rooted at any node and the maximum length of any walk are both specified. For web-scale datasets, only random walks are used because the other neighborhood representations are prohibitively expensive to compute. The set of graph walks thus generated is sequential data, and is used to train a word2vec model to compute node embeddings.
 We note that the random walks used by \citet{DBLP:conf/semweb/RistoskiP16} are uniform random walks, i.e., the edge to traverse next is selected uniformly at random from the set of all outbound edges originating from the current node. Subsequently, \citet*{DBLP:conf/wims/CochezRPP17} have demonstrated that the quality of embeddings and  the accuracy of downstream tasks (such as classification and regression) may be improved by using random walks that are biased as opposed to uniform. We review this approach next. 

\subsubsection{Biased RDF2Vec}
\label{subsubsec:BiasedRDF2Vec}

Biased RDF2Vec formulates a biasing strategy as an edge weighting function, $ weight : E \rightarrow \mathbb{R}^{> 0} $, that assigns a positive real-valued weight to every edge in the graph. Here, $ E $ is the set of graph edges and $ \mathbb{R}^{> 0} $ is the set of positive real numbers. At any node, an outgoing edge is selected with probability proportional to its weight assigned by the weighting function. In other words, if the current node  $v$ has $k$ outgoing edges $e_1, e_2, ..., e_k$, then the probability of selecting edge $e_r$ $(1 \leq r \leq k) $ for the next traversal is given by:
\begin{equation}
P(e_r) = \frac{weight(e_r)}{\sum_{i=1}^k weight(e_i)}
\label{eq:WeightedRandomWalk}
\end{equation}

Various edge weighting functions lead to various biasing strategies, such as, \emph{predicate frequency} where the probability of traversing a (labeled) edge is proportional to the frequency of occurrence of its edge label in the graph, \emph{object frequency} where the probability is proportional to the in-degree of the node to which the edge leads, and so on. We note that the biasing strategies used in Biased RDF2Vec are derived purely from structural information, and do not utilize any knowledge of what the graph really represents. 

In contrast, our approach is to formulate biasing strategies that are aware of the semantics of the underlying graph domains. Such biasing strategies are domain-specific. Our empirical results suggest that in real-world scenarios, embeddings produced using domain-specific strategies may outperform embeddings produced based on structure alone, in terms of accuracy of downstream tasks such as classification.

\subsection{Domain-Aware Biased RDF2Vec}
\label{subsec:DomainAwareBiasedRDF2Vec}

As mentioned in Section \ref{sec:RelatedWork}, we propose an approach to generate node embeddings that uses graph walks biased by an understanding of the underlying semantics. Specifically, we present a framework consisting of the following steps:
\begin{enumerate}
\item \textbf{Data Exploration.} We begin by identifying all the entity types (node types) and relationship types (edge types)\footnote{Entity-sets and relationship-sets in E-R terminology}, and for every relationship type we identify all pairs of entity types that it connects. 
\item \textbf{Attribute Selection.} We examine the \emph{end goal} (e.g., inferring the value of a property of an entity such as a class label) and identify aspects of the graph (e.g., node labels, edge types) that may be relevant to that goal.
\item \textbf{Strategy Definition.} We use the selected attributes to define a family of biasing strategies that increase (or decrease) the likelihood of selecting an edge during any random walk.
\item \textbf{Evaluation.} We generate walks using these biasing strategies and, treating these walks as sentences, apply word2vec to compute node embeddings. The quality of the embeddings is measured using a supervised learning task such as classification of nodes based on their corresponding embeddings. 
\end{enumerate}

We now illustrate this approach using a real-world example.

\subsubsection{The AIFB Dataset}
\label{subsubsec:AIFB}

The AIFB Dataset is an RDF dataset pertaining to the Institute for Applied Informatics and Formal Description Methods at the Karlsruhe Institute of Technology\footnote{\url{http://www.aifb.kit.edu/web/Hauptseite}}. It describes the inter-relationships between persons (e.g., Professors, Students), research topics, projects, publications etc.\ and was used by \citet{DBLP:conf/semweb/BloehdornS07} to predict the research group that any person is affiliated with (see Figure 1 of their paper for a depiction of the entities and relationships in this dataset). We now demonstrate how the steps outlined above can yield useful biasing strategies specific to the AIFB domain. In the following discussion, only the Data Exploration, Attribute Selection, and Strategy Definition steps are described, while the Evaluation steps are presented in Section \ref{sec:Evaluation}.

\paragraph{Strategy 1}

\subparagraph{Data Exploration.}
We begin with the observation that the RDF graph for AIFB, downloaded from  \url{http://data.dws.informatik.uni-mannheim.de/rmlod/LOD_ML_Datasets/data/datasets/RDF_Datasets/AIFB/}, explicitly includes nodes (i.e., entities) of type ``research group'', as well as  edges (i.e., relationships) labeled ``affiliation'' from persons to research groups and edges labeled ``member'' from research groups to persons for the training and test instances. \citet{DBLP:conf/semweb/RistoskiP16} show that  RDF2Vec can learn the equivalence between the  target of the classification problem, namely, research group affiliation, and the concept modeled using the above node and edge types. We note that this approach is different from that of  \citet{DBLP:conf/semweb/BloehdornS07} who treat research group affiliation as a latent variable to be inferred from the graph. We extend \citeauthor{DBLP:conf/semweb/RistoskiP16}'s approach.

\subparagraph{Attribute Selection.}
The random walks used by RDF2Vec do not exploit a priori knowledge of the correspondence between the notion of affiliation and edges labeled ``member'' and ``affiliation'' in the graph; specifically, the walks are agnostic to what these labels \emph{mean}. On the other hand, if a human is shown a (small) subgraph of this data, he/she will probably utilize this correspondence to focus on these edges more than other edges. Our first strategy formalizes this intuition and uses walks that preferentially select edges with these labels for traversal more often than edges with other labels.

\subparagraph{Strategy Definition.}
Recall from Equation \ref{eq:WeightedRandomWalk} that every assignment of weights to edges defines a biased random walk strategy. Based on the above discussion, we assign weight $\hp{w}{high}$ to every edge labeled ``member'' or ``affiliation'', and $\hp{w}{low}$ to every other edge, where hyperparameters $\hp{w}{high}$ and $\hp{w}{low}$ 
satisfy $ 0 < \hp{w}{low} < \hp{w}{high} $.
The resulting strategy is shown in Algorithm  \ref{alg:AIFBBias1}. 
 We note that normalization of weights inside the weighting function is not necessary because the weighted random walk algorithm uses Equation \ref{eq:WeightedRandomWalk} to normalize the weights before choosing the next edge to traverse. We also note that since this strategy relies on hyperparameters, the quality of embeddings and the accuracy of downstream tasks are dependent on hyperparameter tuning. Hyperparameter tuning is necessary for all algorithms presented in this paper.
 
\begin{algorithm}[ht]
\begin{algorithmic}
\If {$type(edge) \in \{member, affiliation\}$}
	\State $ weight \leftarrow \hp{w}{high} $ 
\Else
	\State $ weight \leftarrow \hp{w}{low} $ 
\EndIf
\State \Return $ weight $
\end{algorithmic}
\caption{AIFB-Weight-Function-1 ($ edge, \hp{w}{low}, \hp{w}{high} $)}
\label{alg:AIFBBias1}
\end{algorithm}

While this strategy relies solely on relationship types (i.e., edge labels), a strategy may also rely solely on entity types (i.e., node types), or on both relationship and entity types. The next two strategies uses entity types. 

\paragraph{Strategy 2}

\subparagraph{Data Exploration.} From the AIFB RDF graph, we observe the in addition to persons and research groups, the graph also has nodes representing projects and research topics. Moreover, there are ``is about'' relationships connecting projects and topics, ``works at project'' relationships connecting persons and projects, and ``is worked on by'' relationships connecting research topics and persons\footnote{However, the ``sub-topic'' relationship amongst topics, as depicted in Figure 1 by \citet{DBLP:conf/semweb/BloehdornS07}, is \emph{not} present in the RDF graph.}.

\subparagraph{Attribute Selection.} Based on the intuition that people affiliated with the same research group have a higher likelihood of working on the same project or research topic, we explore whether graph walks that are biased towards nodes of type ``person'', ``project'', ``research topic'' and ``research group'' are helpful in predicting affiliation.

\subparagraph{Strategy Definition.} A strategy based on the above intuition is shown in Algorithm \ref{alg:AIFBBias2}, where any edge that leads to a node of type ``person'', ``project'', ``research group'' or ``research topic'' is assigned weight $\hp{w}{high}$ whereas all other edges are assigned weight $\hp{w}{low}$. Here, hyperparameters $
\hp{w}{high}$ and  $\hp{w}{low}$  satisfy 
$ 0 < \hp{w}{low} < \hp{w}{high} $. Like Algorithm \ref{alg:AIFBBias1}, this algorithm requires parameter tuning.  However, unlike Algorithm \ref{alg:AIFBBias1} that selects edges based on edge labels, this strategy selects edges based on the types of nodes to which those edges lead.

\begin{algorithm}
\begin{algorithmic}
\State $end\_node \leftarrow edge.end\_node $
\If {$type(end\_node) \in \{Person, Project, ResearchGroup, ResearchTopic\}$}
	\State $ weight \leftarrow \hp{w}{high} $ 
\Else
	\State $ weight \leftarrow \hp{w}{low} $ 
\EndIf
\State \Return $ weight $
\end{algorithmic}
\caption{AIFB-Weight-Function-2 ($edge, \hp{w}{low}, \hp{w}{high}$)}
\label{alg:AIFBBias2}
\end{algorithm}

Both Algorithm \ref{alg:AIFBBias1} and Algorithm \ref{alg:AIFBBias2} select some edges more often than the rest; our next algorithm is based on the idea of selecting some edges \emph{less} often than the rest. 

\paragraph{Strategy 3}

\subparagraph{Data Exploration.} 
We observe that the AIFB RDF graph has nodes representing publications, and edges connecting persons and publications to indicate authorship.

\subparagraph{Attribute Selection.}
We begin with the hypothesis that co-authorship in a publication is a  relationship that tends to connect members of the same research group. However, when we experiment with strategies that bias graph walks towards publication nodes, we find that these strategies have a \emph{negative} impact on classification accuracy, which suggests that there may be a substantial number of publications whose authors do not belong to the same research group. In fact, in Section 4.2 of their paper, \citet{DBLP:conf/semweb/BloehdornS07} mention the existence of several authors in the dataset who are external to the AIFB department. As a result, it is difficult to infer affiliation based on co-authorship. This observation leads to the interesting question of whether ignoring co-authorship relationships makes the problem of inferring affiliation simpler. 

\subparagraph{Strategy Definition.} 
To explore this possibility, we devise a strategy to \emph{decrease} the likelihood  of selecting edges that lead to publication nodes. In the strategy shown in Algorithm \ref{alg:AIFBBias3},  any edge leading to a node of type ``publication'' is assigned weight $\hp{w}{low}$ while all other edges are assigned weight $\hp{w}{high}$,  $ 0 < \hp{w}{low} < \hp{w}{high} $.

\begin{algorithm}
\begin{algorithmic}
\State $end\_node \leftarrow edge.end\_node $
\If {$type(end\_node) = Publication$}
	\State $ weight \leftarrow \hp{w}{low} $ 
\Else
	\State $ weight \leftarrow \hp{w}{high} $ 
\EndIf
\State \Return $ weight $
\end{algorithmic}
\caption{AIFB-Weight-Function-3 ($edge, \hp{w}{low}, \hp{w}{high}$)}
\label{alg:AIFBBias3}
\end{algorithm}

We now demonstrate that a strategy may depend both on entity types as well as relationship types, and further that simple biases may be combined to form more complex ones.

\paragraph{Strategy 4}
We combine the approaches of Algorithm \ref{alg:AIFBBias1} (i.e, favoring affiliation and member edges) and Algorithm \ref{alg:AIFBBias3} (i.e., avoiding publication nodes) to get Algorithm \ref{alg:AIFBBias4} which does both. This algorithm has three hyperparameters, $\hp{w}{low}$, $w$, and $\hp{w}{high}$, with $ 0 < \hp{w}{low} < w < \hp{w}{high} $ . Any edge that leads to a node of type ``publication'' is assigned weight $\hp{w}{low}$, any edge whose label is ``affiliation'' or ``member'' is assigned weight $\hp{w}{high}$, and all other edges are assigned weight $w$.

\begin{algorithm}
\begin{algorithmic}
\State $end\_node \leftarrow edge.end\_node $
\If {$type(end\_node) = Publication$}
\State $weight \leftarrow \hp{w}{low} $ 
\Else
\If {$type(edge) \in \{affiliation, member\}$}
\State $weight \leftarrow \hp{w}{high} $ 
\Else
\State $weight \leftarrow w $ 
\EndIf
\EndIf
\State \Return $ weight $
\end{algorithmic}
\caption{AIFB-Weight-Function-4 ($edge, \hp{w}{low}, w, \hp{w}{high}$)}
\label{alg:AIFBBias4}
\end{algorithm}

\paragraph{}
As demonstrated by our experimental results in the next section, these simple domain-aware biasing strategies achieve substantially higher classification accuracy compared to several domain-independent strategies, namely, predicate frequency, inverse predicate frequency, object frequency, inverse object frequency, and uniform random walks. At the same time, the domain-aware strategies presented in this section don't require graph-wide calculations that many domain-independent strategies rely on, such as computing  frequency distributions of edge labels in the graph, or computing in-degrees of all nodes with at least one incoming edge in the graph.

In the next section, we demonstrate similar results for a British Geological Society (BGS) dataset publicly available in RDF format.

\subsubsection{The BGS Dataset}
\label{subsubsec:BGS}

As described by \citet{DBLP:conf/semweb/RistoskiVP16}, the British Geological Society (BGS) RDF dataset documents  observed geological properties of rock types in Great Britain. This dataset was used in machine learning by  \citet{DBLP:conf/pkdd/Vries13} to predict lithogenesis (method of formation) types of named rock units. We now apply our methodology to derive domain-specific biased random walk strategies for the BGS dataset.

\paragraph{Strategy 1}

\subparagraph{Data Exploration.} As with AIFB, we note that the RDF graph for BGS, available from \url{http://data.dws.informatik.uni-mannheim.de/rmlod/LOD_ML_Datasets/data/datasets/RDF_Datasets/BGS/}, contains nodes(i.e., entities) representing lithogenesis types, and 
edges (i.e., relationships) labeled ``hasLithogenesis'' associating rock types to lithogenesis types. This suggests that when trained on the BGS graph, RDF2Vec learns that the equivalence between the target attribute of the classification problem and the concept represented by nodes of type ``lithogenesis'' together with edges labeled ``hasLithogenesis''.

\subparagraph{Attribute Selection.} However, the random walks used by RDF2Vec do not exploit any prior knowledge of this equivalence; in particular, such walks are agnostic to the \emph{meaning} of the node type ``lithogenesis'' and edge label ``hasLithogenesis''. To exploit this semantic information, we explore  walks where edges labeled ``hasLithogenesis'' have a higher likelihood of being traversed compared to other edges.

\subparagraph{Strategy Definition.} Our strategy resulting from the above discussion is shown in Algorithm \ref{alg:BGSBias1}, where any edge labeled ``hasLithogenesis'' is assigned weight $\hp{w}{high}$ whereas other edges are assigned weight $\hp{w}{low}$, with $ 0 < \hp{w}{low} < \hp{w}{high}$

\begin{algorithm}
\begin{algorithmic}
\If {$type(edge) = hasLithogenesis$}
	\State $ weight \leftarrow \hp{w}{high} $ 
\Else
	\State $ weight \leftarrow \hp{w}{low} $  
\EndIf
\State \Return $ weight $
\end{algorithmic}
\caption{BGS-Weight-Function-1 ($edge, \hp{w}{low}, \hp{w}{high}$)}
\label{alg:BGSBias1}
\end{algorithm}

\paragraph{Strategy 2}

In Section \ref{sec:Evaluation}, we show that biasing walks in favor of ``hasLithogenesis'' edges substantially improves classification accuracy. However, it is also interesting to explore the impact of biasing walks against these edges, to gain insight into the usefulness of semantic information contained in the rest of the graph.  
One such biasing strategy is in Algorithm \ref{alg:BGSBias2}.

\begin{algorithm}
\begin{algorithmic}
\If {$type(edge) = hasLithogenesis$}
	\State $ weight \leftarrow \hp{w}{low} $  
\Else
	\State $ weight \leftarrow \hp{w}{high} $   
\EndIf
\State \Return $ weight $
\end{algorithmic}
\caption{BGS-Weight-Function-2 ($edge, \hp{w}{low}, \hp{w}{high}$)}
\label{alg:BGSBias2}
\end{algorithm}
 
 In Algorithm \ref{alg:BGSBias2}, we note that setting $\hp{w}{low}$ to be several orders of magnitude less than $\hp{w}{high}$ effectively amounts to removing the edges labeled ``hasLithogenesis'' and performing uniform random walks on the remainder of the graph. We show in Section \ref{sec:Evaluation} that under the above conditions Algorithm \ref{alg:BGSBias2} achieves approximately the same accuracy as uniform random walks. 
 
An interesting aspect of the BGS dataset is the hierarchical organization of categories. In the next section, we discuss a biasing strategy that utilizes this hierarchy.
 
 \paragraph{Strategy 3}
 
 \subparagraph{Data Exploration.} In the BGS dataset, concepts are organized in a hierarchy using edges labeled ``broader'' and ``narrower''. Intuitively, we expect rock types that are similar in the concept hierarchy to have similar modes of origin, which suggests that the distance between their vector representations should be  small.
 
 \subparagraph{Attribute Selection.} To enable graph walks to discover concept hierarchies, we use a strategy that heavily biases walks towards edges labeled ``broader'' and ``narrower'', while rarely traversing edges labeled ``hasLithogenesis''.
 
 \subparagraph{Strategy Definition.} One implementation of this idea is shown in Algorithm \ref{alg:BGSBias3}, where the hyperparameters satisfy $ 0 < \hp{w}{low} < w < \hp{w}{high} $.

\begin{algorithm}[ht]
\begin{algorithmic}
\If {$type(edge) = hasLithogenesis$}
	\State $ weight \leftarrow \hp{w}{low} $  
\Else
	\If {$type(edge) \in \{broader, narrower\}$}
    	\State $weight \leftarrow \hp{w}{high} $ 
    \Else
		\State $ weight \leftarrow w $  
	\EndIf
\EndIf
\State \Return $ weight $
\end{algorithmic}
\caption{BGS-Weight-Function-3 ($edge, \hp{w}{low}, w, \hp{w}{high}$)}
\label{alg:BGSBias3}
\end{algorithm}

In Section \ref{subsubsec:AIFB}, we had observed that
walks on the AIFB graph that include edges irrelevant to the downstream task can reduce accuracy. This is true for the BGS dataset as well, as we show next. 

\subparagraph{Strategy 4}

We empirically found that edges labeled ``inScheme'' contribute negatively to the accuracy of the classification task. This observation leads to Algorithm \ref{alg:BGSBias4} that tends to avoid these edges.

\begin{algorithm}[ht]
\begin{algorithmic}
\If {$type(edge) = inScheme$}
	\State $ weight \leftarrow \hp{w}{low} $ 
\Else
	\State $ weight \leftarrow \hp{w}{high} $  
\EndIf
\State \Return $ weight $
\end{algorithmic}
\caption{BGS-Weight-Function-4 ($edge, \hp{w}{low}, \hp{w}{high}$)}
\label{alg:BGSBias4}
\end{algorithm}
 
As with AIFB, it is possible to combine simple biasing strategies for walks on the BGS graph into more sophisticated strategies, as shown next.

\paragraph{Strategy 5} Similar to Algorithm \ref{alg:BGSBias3}, the goal here is to enable graph walks to exploit concept hierarchies. However, once a walk has arrived at the most general concept reachable, we want to use the ``hasLithogenesis edge'' (if available) to access the lithogenesis type for that general concept. Intuitively, this strategy is expected to succeed if specialized rock types inherit lithogenesis types from more general rock categories. The strategy is shown in Algorithm \ref{alg:BGSBias5}, where the hyperparameters satisfy $ 0 < \hp{w}{low} < w < \hp{w}{high} $.

\begin{algorithm}[ht]
\begin{algorithmic}
\If {$type(edge) = broader$}
	\State $ weight \leftarrow \hp{w}{high} $ 
\Else
	\If {$type(edge) = hasLithogenesis$}
    	\State $ start\_node = edge.start\_node $
        \If {$start\_node$ has an outgoing edge of type $broader$}
        	\State $ weight \leftarrow \hp{w}{low} $ 
         \Else
            \State $ weight \leftarrow \hp{w}{high} $ 
          \EndIf
    \EndIf
		\State $ weight \leftarrow w $  
\EndIf
\State \Return $ weight $
\end{algorithmic}
\caption{BGS-Weight-Function-5 ($edge, \hp{w}{low}, w, \hp{w}{high}$)}
\label{alg:BGSBias5}
\end{algorithm}

In the next section, we experimentally evaluate our embedding methods and compare them against techniques that use domain-independent biased walks and uniform random walks.

\section{Evaluation}
\label{sec:Evaluation}

For our empirical study, we have implemented a framework for extracting random walks from any RDF graph and subsequently deriving node embeddings from those walks. This framework permits a broad range of biasing strategies including domain-specific, domain-independent, and uniform random walks, making it suitable for experimental evaluation of all three kinds of strategies. This section begins with a description of the framework, followed by experimental results and discussion.

\subsection{Experimental Setup}
\label{subsec:Setup}

\begin{figure}[ht]
\begin{center}
\includegraphics[scale=0.4]{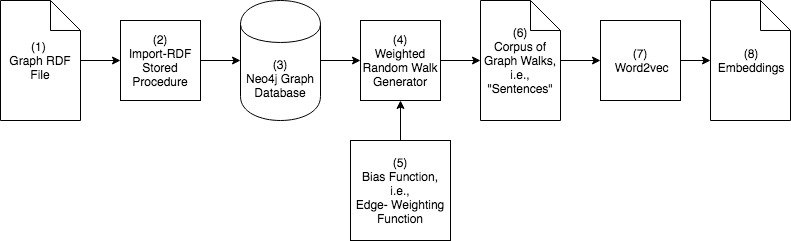}
\caption{Node embedding framework for RDF graphs.}
\label{fig:NodeEmbeddingFramework}
\end{center}
\end{figure}

Our node embedding framework for RDF graphs is depicted in Figure \ref{fig:NodeEmbeddingFramework}
and is composed of the following components:
\begin{enumerate}
\item \textbf{RDF File.} The starting point is an RDF document representing a graph. The RDF documents representing AIFB and BGS are available at URIs provided in Sections \ref{subsubsec:AIFB} and \ref{subsubsec:BGS}, respectively.
\item \textbf{RDF Importer.} We use the neosemantics\footnote{https://github.com/jbarrasa/neosemantics} package, specifically, the \texttt{semantics.importRDF} stored-procedure in that package, to read the RDF file into a graph database.
\item \textbf{Graph Database.} We use Neo4j\footnote{https://neo4j.com/} as our database management system for storing and querying graphs.
\item \textbf{Weighted Random Walk Generator.} This is a program we have written in Python that accepts an edge-weighting function as input, and performs weighted random walks on the graph stored in Neo4j, using Equation \ref{eq:WeightedRandomWalk} for edge selection. The program uses the Py2Neo\footnote{http://py2neo.org} connectivity library to query the Neo4j database, and generates a maximum of $n$ walks rooted at every node of the graph, where $n$ is an input parameter that we set to $100$ in our experiments. Every walk can include one, two, three, or four edge traversals (i.e., hops). 
\item \textbf{Edge-Weighting Function.} Edge-weighting functions are Python implementations of  biasing strategies. Since the Weighted Random Walk Generator accepts this function as input, various types of biased random walks can be easily generated by implementing the corresponding edge-weighting functions, which typically require very few lines of code. 
\item \textbf{Corpus of Graph-Walk ``Sentences''.} The random walks, printed out by the Weighted Random Walk Generator one walk at a time, are typically redirected for storage into a plain text file. This file plays the role of a corpus of sentences.
\item \textbf{Word2vec.} We use the Gensim\footnote{https://radimrehurek.com/gensim/} implementation of word2vec, written by \citet{rehurek_lrec}. Word2vec scans the sentences generated in the previous step to compute embeddings for each word in the sentence (i.e., each node in the graph). Our word2vec parameter settings closely follow those of  \citet{DBLP:conf/semweb/RistoskiP16}. In particular, we set the window size to $5$, number of epochs (i.e., the number of passes through the entire corpus) to $10$, number of negative samples to $25$, and the dimensionality of the embedding space to $200$ and $500$. However, since we haven't observed a substantial improvement in performance at the higher dimensionality, only results for $200$ dimensions are reported below. 
\item \textbf{Embeddings.} Finally, the output of the system is a set of embeddings, i.e., dense vector representations of every node in the graph. The embeddings may be stored as a Python pickle file, or directly used by downstream components such as classifiers.
\end{enumerate}

We evaluate the quality of the resulting embeddings using $k$-Nearest Neighbor ($k$-NN) classification on the AIFB and BGS datasets. While  \citet{DBLP:conf/semweb/RistoskiP16} use $k=3$, we set $k=4$ for AIFB and $k=10$ for BGS so that our implementation can reproduce the baseline results for uniform (i.e., unbiased) random walks reported in their paper. The training and test datasets, available from the same URIs as the RDF files, consist only of node-identifiers and class labels (as well as a row index with respect to the training/test files), where the node-identifiers are the same identifiers as those used in the RDF graphs (and therefore in the Neo4j database and in the extracted graph walks). Notably, no additional tabular attributes are used to guide the classification. We measure classification accuracy as the percentage of correctly classified test instances. 
We next present results obtained from our experiments.

\subsection{Experimental Results}
\label{subsec:Results}

Classification results for the AIFB dataset are presented in Table \ref{tbl:AIFB}. In this table, all numbers correspond to Continuous Bag of Words (CBOW) word2vec embedding into a $200$-dimensional space. Increasing the dimensionality to $500$ does not improve the accuracy appreciably. Skip-gram word2vec reduces our measured accuracy for AIFB.

\begin{table}[ht]
\begin{center}
\begin{tabular}{|r|r|c|}
\hline
\multicolumn{2}{|r|}{\textbf{Biasing Strategy}} & \textbf{CBOW $200$ $(\%)$} \\
\hline
 \multicolumn{2}{|r|}{Uniform} & $68.64$ \\
\hline
\multirow{4}{*}{Domain-independent} & 
  Predicate Frequency & $54.03$ \\
  & Inverse Predicate Frequency & $71.33$ \\
  & Object Frequency & $77.42$ \\
  & Inverse Object Frequency & $50.56$ \\
\hline
\multirow{4}{*}{Domain-specific} &
  AIFB-Weight-Function-1 & $88.28$ \\
  & AIFB-Weight-Function-2 & $89.94$ \\
  & AIFB-Weight-Function-3 & $91.56$ \\
  & AIFB-Weight-Function-4 & $99.86$ \\
\hline
\end{tabular}
\caption{$k$-Nearest Neighbor classification ($k=4$) accuracy for the AIFB dataset.}
\label{tbl:AIFB}
\end{center}
\end{table}

As mentioned in the previous section, hyperparameter setting affects the quality of embeddings. In this study, we have tuned hyperparameters manually; use of automated hyperparameter tuning is a possible extension of this work. Hyperparameter values used in the domain-specific biasing strategies in Table \ref{tbl:AIFB} are shown Table \ref{tbl:AIFBHyperparameters}.

\begin{table}[ht]
    \begin{center}
        \begin{tabular}{|c|c|}
            \hline
            {\bf Algorithm} & {\bf Hyperparameters} \\ \hline
            AIFB-Weight-Function-1 & 
            $ \hp{w}{low} = 0.1, \hp{w}{high} = 10 $ \\ \hline
            AIFB-Weight-Function-2 &
            $ \hp{w}{low} = 0.1, \hp{w}{high} = 10   $\\ \hline
            AIFB-Weight-Function-3 &
            $ \hp{w}{low} = 0.1, \hp{w}{high}  = 10 $\\ \hline
            AIFB-Weight-Function-4 &
            $ \hp{w}{low} = 0.1, w = 10, \hp{w}{high} = 100 $  \\ \hline
        \end{tabular}
    \end{center}
    \caption{Hyperparameter values used in the domain-specific biasing strategies for the AIFB dataset.}
    \label{tbl:AIFBHyperparameters}
\end{table}

Classification results for the BGS dataset are presented in Table \ref{tbl:BGS}. In this table, we include results for Continuous Bag of Words (CBOW) as well as Skip-gram word2vec embeddings into a $200$-dimensional space. For the BGS dataset, we include Skip-gram results since in some cases, Skip-gram provides slightly better results than CBOW. Increasing the dimensionality to $500$ does not improve the accuracy appreciably.

\begin{table}[ht]
\begin{center}
\begin{tabular}{|r|r|c|c|}
\hline
\multicolumn{2}{|r|}{\textbf{Biasing Strategy}} & \textbf{\makecell{CBOW \\ $200$ $(\%)$}} & \textbf{\makecell{Skipgram \\ $200$ $(\%)$}} \\
\hline
 \multicolumn{2}{|r|}{Uniform} & $74.45$ & $74.10$ \\
\hline
\multirow{4}{*}{Domain-independent} & 
  Predicate Frequency & $59.52$ & $72.41$ \\
  & Inverse Predicate Frequency & $88.45$ & $93.10$ \\
  & Object Frequency & $21.52$ & $34.48$ \\
  & Inverse Object Frequency & $50.79$ & $34.48$ \\
\hline
\multirow{5}{*}{Domain-specific} &
  BGS-Weight-Function-1 & $84.21$ & $90.07$ \\
  & BGS-Weight-Function-2 & $75.86$ & $75.86$ \\
  & BGS-Weight-Function-3 & $65.52$ & $68.97$ \\
  & BGS-Weight-Function-4 & $70.14$ & $73.76$ \\
  & BGS-Weight-Function-5 & $80.14$ & $78.31$ \\
\hline
\end{tabular}
\caption{$k$-Nearest Neighbor classification ($k=10$) accuracy for the BGS dataset.}
\label{tbl:BGS}
\end{center}
\end{table}

Hyperparameter values used in the domain-specific biasing strategies in Table \ref{tbl:BGS} are shown in Table \ref{tbl:BGSHyperparameters}.

\begin{table}[ht]
    \begin{center}
        \begin{tabular}{|c|c|}
            \hline
            {\bf Algorithm} & {\bf Hyperparameters} \\ \hline
            BGS-Weight-Function-1 & 
            $  \hp{w}{low} = 0.1, \hp{w}{high} = 10 $ \\ \hline
            BGS-Weight-Function-2 &
            $ \hp{w}{low} = 0.001, \hp{w}{high} = 1   $\\ \hline
            BGS-Weight-Function-3 &
            $ \hp{w}{low} = 0.1, w = 10, \hp{w}{high} = 100  $\\ \hline
            BGS-Weight-Function-4 &
            $ \hp{w}{low} = 0.1, \hp{w}{high} = 1 $  \\ \hline
        \end{tabular}
    \end{center}
    \caption{Hyperparameter values used in the domain-specific biasing strategies for the BGS dataset.}
    \label{tbl:BGSHyperparameters}
\end{table}

\subsection{Remarks}
\label{subsec:Remarks}

In the case of AIFB, we observe that domain-specific biases provide a substantial improvement in classification accuracy compared to domain-independent ones. In particular, the simple strategy of avoiding publication nodes (AIFB-Weight-Function-3) leads to a classification accuracy of $91.56\%$. When we avoid publication nodes and also preferentially select affiliation and member edges for traversal (AIFB-Weight-Function-4), the classification accuracy increases to $99.86\%$. We further note that the domain-specific algorithms do not require graph-wide calculations that the domain-independent biased walks rely on, such as the frequency distribution of edge labels (i.e., predicate frequencies).  In the domain-specific walks, all decisions are made on locally available data which, due to locality of reference, is also highly cacheable.

In the case of BGS, we observe that the most successful domain-specific biasing strategy (BGS-Weight-Function-1) performs in the ballpark of the most successful domain-independent biasing strategy (inverse predicate frequency). Even under these circumstances, the domain-specific BGS-Weight-Function-1 provides the advantage of not requiring graph-wide computation of predicate (i.e., edge-label) frequencies. We haven't yet been able to define a domain-specific strategy that can outperform the best domain-independent strategy. 

Of the two real-world domains studied here, the AIFB domain (involving researchers, research groups, publications, etc.) is closer to our own field of work than the BGS domain (involving rock formations) which pertains to geology. As a result, it was easier for us to effectively exploit AIFB semantic information to devise biasing strategies that outperform domain-independent walks, whereas in BGS, our strategies could match the performance of domain-independent walks but not outperform them. It is possible that a domain expert in the methods of rock formation might be able to invent biasing strategies for BGS that outperform domain-independent methods.

\section{Conclusion}
\label{sec:Conclusion}

To summarize, we have demonstrated that semantics of underlying graph domains provide valuable information for computing graph embeddings. We have presented a framework for devising biased random walk strategies that are informed by what the nodes and edges really mean, and use knowledge of downstream machine learning tasks to identify which graph substructures are more relevant, and therefore should be included more often in graph walks, than others . We have used this framework in two real-life domains, and shown that the resulting embeddings are simple to implement and achieve equal or higher accuracy in machine learning tasks compared to domain independent approaches.  Our results also suggest that when this approach is applied to other real-life domains in future, domain expertise may be a key ingredient in the development of biasing algorithms that outperform domain-independent methods.

\bibliographystyle{plainnat}
\bibliography{mybib.bib}
\end{document}